\documentclass{article}

\usepackage{arxiv}

\usepackage[utf8]{inputenc} 
\usepackage[T1]{fontenc}    
\usepackage{hyperref}
\usepackage{url}            
\usepackage{booktabs}       
\usepackage{amsfonts}       
\usepackage{nicefrac}       
\usepackage{microtype}      
\usepackage{lipsum}
\usepackage{graphicx}
\usepackage{lipsum}
\usepackage{float} 
\usepackage{enumitem}
\graphicspath{ {./images/} }

\title{BIDWESH: A Bangla Regional Based Hate Speech Detection dataset}

\author{
Azizul Hakim Fayaz \\
  Department of Computer Science and Engineering\\
  Southeast University\\
  Dhaka, Bangladesh\\
  \texttt{azizulhakimfayaz@gmail.com} \\
   \And
 MD. Shorif Uddin \\
  Department of Computer Science and Engineering\\
  Southeast University\\
  Dhaka, Bangladesh \\
  \texttt{mdshorifuddinbappu@gmail.com} \\
  \And
 Rayhan Uddin Bhuiyan \\
  Department of Computer Science and Engineering\\
  Southeast University\\
  Dhaka, Bangladesh \\
  \texttt{rayhan.scholar@gmail.com} \\
  \And
 Zakia Sultana \\
  Department of Computer Science and Engineering\\
  Southeast University\\
  Dhaka, Bangladesh\\
  \texttt{zsultana855@gmail.com} \\
  \And
 Md. Samiul Islam  \\
  Department of Computer Science and Engineering\\
  Southeast University\\
  Dhaka, BangladeshDhaka, Bangladesh \\
  \texttt{samiulislamsawon09@gmail.com} \\
  \And
 Bidyarthi Paul \\
  Department of Computer Science and Engineering\\
  Southeast University\\
  Dhaka, Bangladesh \\
  \texttt{bidyarthi.paul@seu.edu.bd} \\
  \And
 Tashreef Muhammad \\
  Department of Computer Science and Engineering\\
  Southeast University\\
  Dhaka, Bangladesh\\
  \texttt{tashreef.muhammad@seu.edu.bd}\\
  \And
 Shahriar Manzoor \\
  Department of Computer Science and Engineering\\
  Southeast University\\
  Dhaka, Bangladesh\\
  \texttt{smanzoor@seu.edu.bd}\\
}

\def\bng{\bngx}

\font\bngx=bang10

\def\*#1*#2{o\null{#2}{#1}}

\def\sh#1{\setbox0=\hbox{#1}%
     \kern-.02em\copy0\kern-\wd0
     \kern.04em\copy0\kern-\wd0
     \kern-.02em\raise.0433em\box0 }
\begin{document}
\maketitle
\begin{abstract}

Hate speech on digital platforms has become a growing concern globally, especially in linguistically diverse countries like Bangladesh, where regional dialects play a major role in everyday communication. Despite progress in hate speech detection for standard Bangla, Existing datasets and systems fail to address the informal and culturally rich expressions found in dialects such as Barishal, Noakhali, and Chittagong. This oversight results in limited detection capability and biased moderation, leaving large sections of harmful content unaccounted for. To address this gap, this study introduces BIDWESH, the first multi-dialectal Bangla hate speech dataset, constructed by translating and annotating 9,183 instances from the BD-SHS corpus into three major regional dialects. Each entry was manually verified and labeled for hate presence, type (slander, gender, religion, call to violence), and target group (individual, male, female, group), ensuring linguistic and contextual accuracy. The resulting dataset provides a linguistically rich, balanced, and inclusive resource for advancing hate speech detection in Bangla. BIDWESH lays the groundwork for the development of dialect-sensitive NLP tools and contributes significantly to equitable and context-aware content moderation in low-resource language settings.


\end{abstract}



\section{Introduction}
In recent years, the exponential growth of online communication has led to a concerning rise in hate speech across social media platforms, forums, and digital communities. Hate speech, characterized by abusive or derogatory language targeting individuals or groups based on race, religion, gender, or ethnicity, poses serious threats to mental health, social cohesion, and democratic discourse. Given the vast volume of user-generated content, manual moderation is no longer practical, prompting the need for automated solutions powered by Natural Language Processing (NLP). 

Researchers have focused on creating annotated datasets, developing machine learning models, and improving classification methods. Notable contributions include Davidson et al.(2017)\cite{davidson2017automated}, who introduced an English hate speech dataset and applied logistic regression and SVM, and Badjatiya et al.(2017)\cite{badjatiya2017deep}, who utilized deep learning approaches such as LSTM and gradient boosting for Twitter-based hate speech detection. While these methods have proven effective in high-resource languages, low-resource languages such as Bangla remain underexplored. Earlier research primarily focused on standard Bangla, overlooking the rich variety of regional dialects spoken throughout the country. 

To address linguistic diversity in Bangla NLP, it's important to note that there has not been much work done before in regional dialects. To fulfill this gap, three paper corpora were proposed. First, Vashantor (Faria et al., 2023)\cite{faria2023vashantor} emerged as the pioneering work that made regional dialects publicly available, establishing a foundational resource for dialectal research. Subsequently, Paul et al. (2025)\cite{paul2025ancholik} proposed a hybrid translated regional dialect dataset alongside their NER dataset, which contributes substantially to regional dialect research by combining dialectal NER with dialect-to-standard Bangla translation to support dialect normalization. This hybrid approach significantly advances regional dialect processing by enabling both entity recognition and standardization within dialectal contexts, marking a considerable contribution to the field. Finally, Kundu et al. (2025)\cite{kundu2025anubhuti} proposed ANUBHUTI, a comprehensive emotion detection dataset containing 2,000 dialectal sentences annotated with both thematic (Political, Religious, Neutral) and emotional (Anger, Fear, Sadness, etc.) labels. ANUBHUTI fills a significant gap in sentiment and emotion analysis for Bangla dialects, providing researchers with essential resources for understanding emotional expressions across different regional linguistic varieties. 

\begin{figure}[H]
    \centering
    \includegraphics[width=0.9\textwidth]{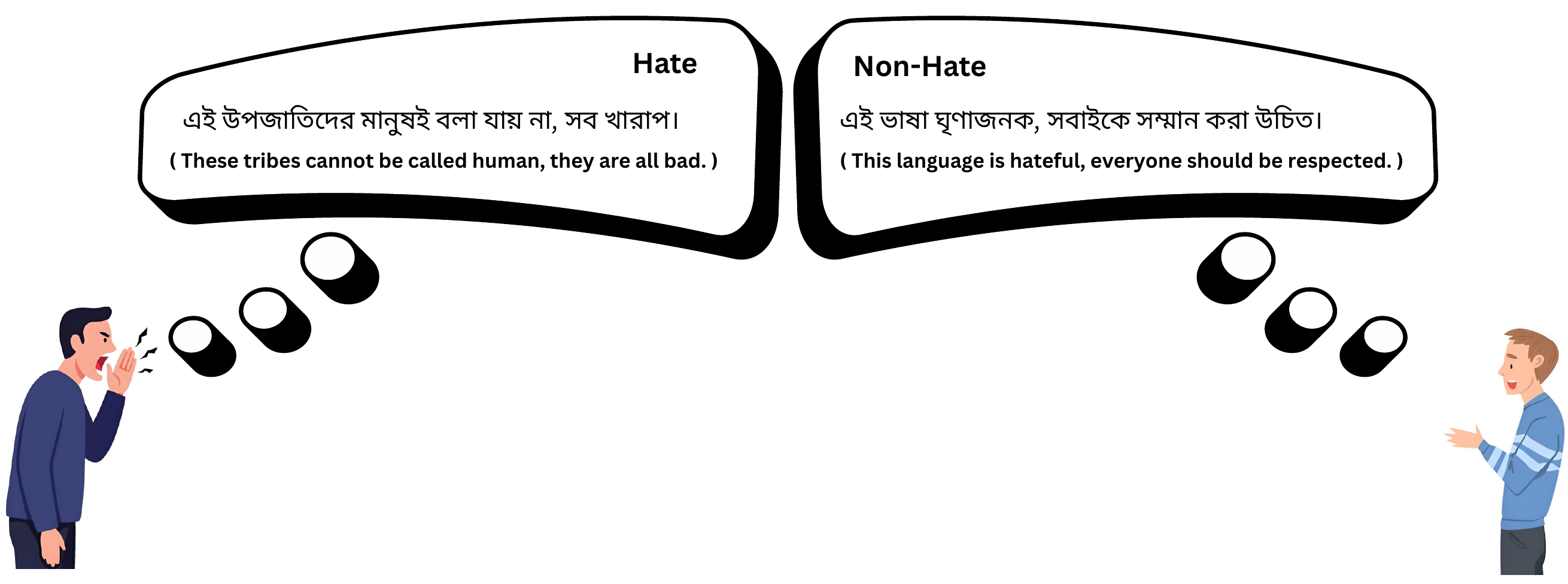}
    \caption{Examples of hate and non-hate sentences across Bangla Language}

    \label{fig:bangla-dialects}
\end{figure}

However, despite these advancements, no existing work has yet addressed hate speech detection in Bangla regional dialects—leaving a critical gap in safeguarding online spaces against toxic content expressed through informal and dialectal language. This study aims to fill that void by introducing a comprehensive dataset and dialect-aware framework for hate speech detection in Bangla, ensuring broader linguistic inclusivity and real-world applicability. Figure~\ref{fig:bangla-dialects} illustrates an example of hate and non-hate sentences in Bangla.

The ability to detect and mitigate hate speech has become increasingly critical as harmful and abusive content can rapidly proliferate across social media platforms, often inciting violence, discrimination, and psychological harm to vulnerable populations. While modern natural language processing (NLP) has achieved remarkable advances in machine translation, text summarization, and conversational AI systems, the ethical imperative of ensuring safe and inclusive digital communication remains a fundamental challenge that extends beyond technical considerations to encompass broader societal responsibilities. Hate speech detection represents not merely a computational problem but a societal necessity essential for upholding democratic values, protecting marginalised communities, and fostering respectful online discourse. However, existing detection systems predominantly rely on training data from formal, high-resource languages and fail to adequately account for the informal, code-switched, and dialect-rich linguistic patterns that characterise authentic online communication. In linguistically diverse nations such as Bangladesh, regional dialects constitute an integral component of everyday digital communication, encompassing expressions of emotion, opinion, and unfortunately, hate speech that often employs local idioms, colloquialisms, and informal grammatical structures. The absence of dialect-aware models results in systematic under-detection of harmful content, particularly when hate speech is deliberately obfuscated through culturally specific language patterns, creating significant gaps in content moderation capabilities. Consequently, developing robust hate speech detection systems capable of understanding and processing regional dialectal variations has evolved from an optional enhancement to an essential requirement for building equitable, secure, and contextually aware NLP systems that authentically reflect the complete spectrum of linguistic diversity inherent in contemporary digital communication. benefit governments and platforms in moderating content. However, if models focus only on standard language, they will miss regional slurs. Prior work on Bangla dialect translation highlights that dialects such as Barishal, Noakhali, and Chittagong were chosen because of their “notable differences” from standard Bangla that lead to communication challenges. Addressing these dialects ensures that hate-speech tools are fair and effective across all Bangla-speaking communities.

In this paper, we introduce a newly constructed multi-dialect Bangla hate speech dataset, developed by adapting the BD-SHS benchmark corpus — the largest publicly available Bangla hate speech dataset with over 50,000 labeled social media comments. Each entry in BD-SHS has been carefully translated and rephrased into the Barishal, Noakhali, and Chittagong dialects, resulting in a parallel corpus that reflects regional linguistic variations while preserving the original semantic and offensive content. This process is guided by linguistic experts and dialectal resources to ensure both accuracy and authenticity.

Our contributions in this work encompass the following key elements:
\begin{itemize}
    \item We propose a multi-dialect Bangla hate speech dataset by adapting BD-SHS into three regional dialects: Barishal, Noakhali, and Chittagong, creating a parallel corpus that supports dialect-aware analysis.
    \item We ensure high-quality translation and contextual integrity through a linguistically informed annotation strategy, preserving both the meaning and hate speech context of the original data.
    \item We leverage dialectal adaptation techniques, inspired by recent advancements, to maintain natural language use within each regional variation.
\end{itemize}

\section{Related works}
While existing works have significantly advanced hate speech detection in standard Bangla, they overlook the complexities of regional dialects, which often carry distinct cultural and linguistic nuances. These dialects introduce challenges such as inconsistent orthography, unique vocabulary, and informal grammar structures. Our study builds upon this foundation by shifting focus to hate speech expressed in underrepresented Bangla dialects. This section explores prior contributions in Bangla, highlighting key datasets like BD-SHS and models such as BanglaHateBERT and Bi-LSTM. It also reviews multilingual efforts  that address hate speech across diverse linguistic landscapes using ensemble and attention-based models. Finally, we examine the emergence of Large Language Models, such as GPT-4 and BanglaBERT, which demonstrate promising capabilities in low-resource and code-switched contexts. Together, these studies establish the groundwork for our dialect-aware approach.

\subsection{Bangla Hate Speech Detection}
The progression of Bangla hate speech detection has advanced significantly, driven by the availability of well-curated datasets and innovative modeling techniques. 

Romim et al.(2022)\cite{romim2022bd} introduced BD-SHS, a large, manually labeled dataset of 50,281 Bangla comments, with 24,156 identified as hate speech, collected from online social media and streaming sites covering diverse social contexts. The paper benchmarks different NLP models, including Support Vector Machine (SVM) and Bidirectional Long Short Term Memory (Bi-LSTM), using linguistic features such as TF-IDF scores and various word embeddings. The Bi-LSTM model trained with informal text embeddings (IFT) achieved the highest F1-score of 91.0\% for hate speech identification, demonstrating superior performance over pre-trained embeddings on formal texts. This work established a foundational benchmark for both binary and multi-class classification. Additionally, their evaluations showed that mBERT outperformed other models in overall performance. 

Jahan et al.(2022) \cite{jahan2022banglahatebert} further advanced the field by developing BanglaHateBERT, introduced BanglaHateBERT, a retrained BERT model for abusive language detection in Bengali. They developed a new 15K manually labeled, balanced dataset of Bengali hate speech from YouTube and Facebook comments, and retrained the model with a large-scale corpus of 1.5 million offensive posts. Benchmarking against generic BERT models (BanglaBERT, IndicBERT, mBERT) and CNNs, BanglaHateBERT consistently outperformed all alternatives. The model achieved the highest accuracy of 94.3\% and an F1 score of 94.1\% on their newly collected balanced dataset. Subsequently, Belal et al.(2023)\cite{belal2023interpretable} introduced a deep learning pipeline for Bengali toxic comment classification, utilizing a new, manually labeled dataset comprising 16,073 instances, with 8,488 identified as toxic and multi-labeled into six categories. Their approach involved a two-stage process. For binary classification, LSTM with BERT Embedding achieved the highest accuracy of 89.42\%.. For multi-label classification, a CNN-BiLSTM with attention mechanism achieved the highest accuracy of 78.92\% and a 0.86 weighted F1-score, outperforming other benchmarked models like Bangla BERT fine-tune. The models' predictions were interpreted using LIME. As the research matured, new datasets and models emerged. 

Haider et al.(2024)\cite{haider2024banth} made a significant contribution by introducing BANTH, the first multi-label dataset for transliterated Bangla hate speech. Along with Bangla-specific and multilingual transformer models, they evaluated large language models such as GPT-3.5 and GPT-4 using prompting strategies including zero-shot, few-shot, and explanation-based approaches. The most notable performance was delivered by GPT-4 combined with Bangla translation and explanation prompting, achieving the highest macro-F1 score among all methods. Concurrently, efforts to enhance model robustness and diversity were undertaken. Paul et al.(2025)\cite{paul2025analyzing} published their work on analyzing emotions in Bangla social media comments using the EmoNoBa dataset, which contains 22,698 Bangla comments labeled for six emotion categories. Their study employed various machine learning models, including LinearSVM, KNN, Random Forest, and Decision Tree with AdaBoost, achieving the best performance with a Decision Tree enhanced by AdaBoost (F1-score of 0.7860). They also utilized LIME for model interpretability, contributing to the broader understanding of emotion analysis in Bengali text and providing valuable resources for affective computing. Similarly, Kundu et al.(2025)\cite{kundu2025anubhuti} also worked in emotion detection but specifically focused on regional dialects, proposing a dataset that addresses emotion analysis in dialectal Bangla, thereby extending the scope of emotion detection research to include regional linguistic variations in hate speech detection systems

Additionally, Paul et al.(2024)\cite{paul2024improving} conducted research on improving Bangla regional dialect detection using BERT, LLMs, and Explainable AI (XAI). Working with the VASHANTOR dataset containing 12,505 speech samples from five regional dialects (Mymensingh, Chittagong, Barishal, Noakhali, and Sylhet), they fine-tuned Bangla BERT and mBERT models while also testing large language models like GPT-3.5 Turbo and Gemini 1.5 Pro. Their approach achieved a maximum accuracy of 88.74\% with Bangla BERT, while GPT-3.5 Turbo demonstrated 64\% accuracy using few-shot learning. The integration of LIME for explainability enhanced model transparency, making their work valuable for understanding regional linguistic variations that could impact hate speech detection across different Bengali dialects.

Mridha et al.(2021)\cite{mridha2021boost} introduced L-Boost, a novel ensemble method for identifying offensive Bengali and Banglish texts from social media. They compiled a new, manually labeled real-life dataset of 16,800 posts and comments34. The study benchmarked various baseline classifiers, including SVM, DT, RF, and LSTM, as well as BERT-based models. Their proposed L-Boost model, combining BERT embedding with LSTM and AdaBoost, achieved the highest accuracy of 95.11\%, significantly outperforming other approaches like Bangla BERT and DeepHateExplainer.Tarin et al.(2025)\cite{tarin2025bengali} proposed an ensemble-based machine learning model for Bengali hate speech detection1. They utilized the BD-SHS dataset, comprising 50,281 manually labeled Bengali comments from social media and streaming platforms, addressing binary classification, multi-label categorization, and target identification tasks. Their approach involved a stacking ensemble model integrating various machine learning classifiers. This stacking ensemble model achieved the highest accuracy of 94.37\% with an F1-score of 94.37\% on the balanced dataset for hate speech identification . 

Keya et al.(2023)\cite{keya2023g} introduced G-BERT, a novel model combining BERT and GRU for hate speech detection in Bengali social media texts. They compiled a new, manually labeled real-life dataset of 20,000 posts, comments, and memes, with nearly half identified as offensive78. G-BERT was benchmarked against various models, including Bangla BERT, LSTM-BERT, AdaBoost-BERT, and L-BOOST9. The proposed G-BERT model consistently outperformed all others, achieving the highest accuracy of 95.56\%. In parallel, Swarnali et al.(2024)\cite{swarnali2024bengali} advanced Bengali text classification using Token-level Adversarial Contrastive Training (TACT) and Label-aware Contrastive (LCL) loss. They utilized datasets including Rokomari Book Review (1,445 instances), Bengali Hate Speech (BHS-M with 5,698 instances, BHS-B with 30,000), and a new custom Daraz Product Review (DPR) dataset (4,004 samples). TACT, applied with BERT embeddings, consistently outperformed previous contrastive learning approaches14. For binary classification, TACT achieved the highest F1-score of 98\% on the RBR dataset, and for multi-class classification, it matched the current benchmark with an F1-score of 91\% on the BHS-M dataset. The significance of cross-lingual and cultural adaptability was also emphasized. 

Ghosh et al.(2022)\cite{ghosh2022hate} compared monolingual and multilingual transformers such as BanglaBERT and XLM-R across Bangla, Hindi, and Assamese, achieving an F1-score of 0.9035, thereby showcasing the potential of multilingual transfer learning. Simultaneously, saha et al.(2024)\cite{saha2024bengalihatecb} introduced BengaliHateCB, a hybrid model trained on both Bengali and Banglish datasets, enhancing performance across linguistic variations.

Finally, explainability and media-aware learning gained prominence. Karim et al.(2021)\cite{karim2021deephateexplainer} presented DeepHateExplainer, which integrates XLM-RoBERTa with explainable AI techniques to interpret predictions on Bangla hate speech, ensuring transparency without compromising accuracy. Additionally Karim et al.(2021)\cite{karim2022multimodal} addressed visual hate content by combining text and image inputs in memes, employing DenseNet and Conv-LSTM models to achieve an F1-score of 0.83, thereby introducing multimodal hate speech detection to the Bangla NLP domain.

\subsection{Hate Speech Detection in Multilingual Languages}
Within the extensive digital landscape, English hate speech detection has experienced numerous milestones, propelled by the intricate nature of language and its varied manifestations online. Researchers have diligently integrated diverse models and methodologies to identify hateful expressions, regardless of their subtlety or veiling. 

Sharma et al.(2025)\cite{sharma2025stop} developed a unique multilingual ensemble that combined the strengths of LSTM, mBERT, and XLM-RoBERTa, akin to weaving diverse threads into a vibrant tapestry. Their model was evaluated on the HopeEDI dataset, which comprises a broad range of hateful and hopeful messages, and achieved an impressive F1-score of 0.93, illustrating how multiple perspectives can collaboratively enhance the understanding of human expression . Subsequently, Sharif et al.(2024)\cite{10439156} examined over 450,000 social media comments from 18 platforms. Their BiLSTM-attention model functioned like a perceptive listener, learning to focus on the most informative words amid noisy dialogues, and attained a notable F1-score of 0.92. This study underscored how attention mechanisms enable machines to comprehend the dynamics of human conversation, even under chaotic conditions. Addressing the challenge of code-mixed languages such as Hinglish, where English and Hindi coexist,

Biradar et al.(2024)\cite{biradar2021hate} developed the TIF-DNN model. By employing translation and fusion techniques, they adeptly uncovered hateful content concealed within blended languages. Their model achieved an F1-score of 0.73, marking an important advancement in managing the complex multilingual communication prevalent in South Asia’s digital spaces. On the other hand, Kodali et al.(2025)\cite{kodali-etal-2025-bytesizedllm} focused on Devanagari-script languages, including Hindi and Marathi. Their customized Attention BiLSTM combined with XLM-RoBERTa embeddings functioned as a cultural interpreter, detecting hate speech as well as identifying its targets, achieving F1-scores of 0.7481 and 0.6715, respectively. This dual insight enriched the comprehension of hate speech beyond mere categorization. 

Aliyu et al.(2022)\cite{aliyu2022herdphobia} illuminated marginalized voices by concentrating on Nigerian languages, Hausa and Nigerian-Pidgin. Utilizing their HERDPhobia dataset and XLM-T model, they attained an exceptional weighted F1-score of 99.83\%, demonstrating that even underrepresented languages can benefit from advanced transformer models supported by dedicated datasets. Additionaly, Tonneau et al.(2024)\cite{tonneau2024languages} delivered a crucial reminder: language is inseparable from culture and geography. They showed that hate speech detection models struggle when applied to data from unfamiliar regions, highlighting the necessity of designing systems that embrace cultural diversity and fairness. Absent this, even the most sophisticated models may falter outside their original training environments.Aligned with this vision,

ghosh et al.(2025)\cite{ghosh2025hate} and Singh et al.(2024)\cite{singh-thakur-2024-generalizable} investigated how transformers and federated learning can bridge linguistic divides while safeguarding user privacy. Ghosh’s research on low-resource Indian languages revealed how multilingual transformers generalize knowledge across languages. Singh’s federated learning approach enabled models to collaborate without sharing raw data, a vital advancement for privacy-preserving hate speech detection across linguistically diverse contexts.

\begin{table}[h]
\centering
\caption{Summary of Bengali Hate Speech Detection Papers}
\renewcommand{\arraystretch}{1.3}
\begin{tabular}{|p{3.2cm}|p{2.5cm}|p{5.3cm}|p{4cm}|}
\hline
\textbf{Author} & \textbf{Dataset} & \textbf{Approach} & \textbf{Model and F1 / Accuracy} \\
\hline
Haider et al.\cite{haider2024banth} & BanTH & BanglaBERT, TB-BERT, GPT-4 (Prompt-based) & GPT-4 + Translation + Exp: Best Macro-F1 \\
\hline
Faria et al.\cite{faria2024investigating} & Hasoc2021, BanTH & BERT, ELECTRA, GPT-3.5 Turbo & GPT-3.5 Turbo: Accuracy 98.53\% \\
\hline
Ghosh et al.\cite{ghosh2022hate} & HASOC, HS-Bangla & XLM-RoBERTa, BanglaBERT & BanglaBERT: F1 0.9035 \\
\hline
Swarnali et al.\cite{swarnali2024bengali} & RBR, BHS-M, DPR & BanglaBERT + TACT & BanglaBERT+TACT: F1 0.98 \\
\hline
Das et al.\cite{das2022hate} & Bengali\_Hate & XLM-RoBERTa, m-BERT, XLM-R, IndicBERT, MuRIL & XLM-RoBERTa \\
\hline
Jahan et al.\cite{jahan2022banglahatebert} & BanglaHateBERT & CNN + FastText, BanglaBERT & BanglaHateBERT: Accuracy 93.1\% \\
\hline
Karim et al.\cite{karim2022multimodal} & Hate Speech v2.0 & XLM-RoBERTa & XLM-RoBERTa: F1 87\% \\
\hline
Karim et al.\cite{karim2021deephateexplainer} & Memes Dataset & Conv-LSTM, ResNet, XLM-R & XLM-RoBERTa: F1 0.83 \\
\hline
Romim et al.\cite{romim2022bd} & BD-SHS & mBERT, BiLSTM, RF & mBERT: F1 0.79 \\
\hline
\end{tabular}
\label{tab:bangla_hate_table}
\end{table}

\subsection{Large Language Models (LLMs) for Hate Speech Detection}
The emergence of Large Language Models has heralded a new era in combating hate speech, providing tools capable of understanding language with remarkable depth and flexibility.

Haider et al.(2024)\cite{haider2024banth} spearheaded this movement in the Bangla domain by introducing the BANTH dataset—a valuable repository of transliterated Bangla hate speech. They investigated how GPT-3.5 and GPT-4 could be prompted with translation and explanatory context, transforming machines from blunt classifiers into insightful analysts. GPT-4, particularly when guided with detailed explanation prompting, distinguished itself by achieving the highest macro-F1 score, setting new standards for intelligent language comprehension in low-resource environments .Expanding on this groundwork, 

Faria et al.(2024)\cite{faria2024investigating} compared conventional transformer models against these powerful LLMs. They found that GPT-3.5 Turbo, employing zero-shot prompting and balanced loss strategies, achieved a remarkable accuracy of 98.53\%. This result not only underscored the immense capabilities of LLMs but also inspired optimism for their application in languages with limited annotated data. Zahid et al.(2025)\cite{zahid2025evaluation} extended this exploration by evaluating LLMs on multilingual datasets covering five global regions. Among candidates such as CodeLlama, Llama2, and DeepSeekCoder, CodeLlama stood out with an F1-score of 52.18\%. Their study emphasized that while LLMs are potent, their effectiveness varies greatly depending on regional and cultural contexts, underscoring the importance of nuanced, location-aware deployment . 

Sharif et al.(2024)\cite{10439156} adopted a hybrid methodology, combining LLMs with traditional BiLSTM models to leverage the strengths of both. This fusion enhanced performance on noisy, large-scale datasets, illustrating that integrated approaches can surpass individual model capabilities. 

Tonneau et al.(2024)\cite{tonneau2024languages} once again stressed the importance of fairness and cultural inclusivity. They advocated for evaluation frameworks that extend beyond accuracy metrics to ensure equitable treatment of diverse populations—an essential consideration as society entrusts machines with increasingly critical social functions .In summary, the evolution of LLMs in hate speech detection is ongoing, filled with both promise and challenges. These models bring power, explainability, and adaptability, yet they also demand vigilance to mitigate bias, uphold ethical standards, and responsibly serve all communities.

Although extensive research has been carried out on hate speech detection in major languages like English, there is a noticeable lack of work focused on regional languages, especially Bangla. Most existing studies and datasets do not address the unique challenges posed by hate speech in Bangla regional contexts, which often include local dialects, cultural references, and code-mixed language. We identified this gap and found that there has been little to no effort specifically targeting Bangla regional hate speech. Therefore, our work aims to fill this research gap by building a relevant dataset and developing models to detect hate speech effectively in Bangla regional content.Table~\ref{tab:bangla_hate_table}, illustrates the Summary of Bengali Hate Speech Detection Papers.

\section{Data Description}

Hate speech detection, a critical subtask of Natural Language Processing (NLP), involves identifying and categorizing offensive, abusive, or toxic language in textual content. It plays a central role in digital safety, particularly in applications such as content moderation, social media monitoring, and online harm reduction. While notable progress has been made in high-resource languages like English due to the availability of large annotated corpora, the same cannot be said for low-resource languages like Bangla. Although Bangla is spoken by millions, the development of hate speech detection tools in this language remains limited. Existing datasets primarily focus on Standard Bangla, overlooking the linguistic richness of regional dialects such as Chittagong, Barishal, and Noakhali. These dialects carry distinct phonological, lexical, and syntactic variations that significantly influence how hate speech is articulated. Moreover, dialectal expressions are often more emotionally charged and contextually grounded, offering deeper insight into region-specific offensive discourse—especially in politically and religiously sensitive contexts.

\begin{figure}[H]
    \centering
    \includegraphics[width=0.9\textwidth]{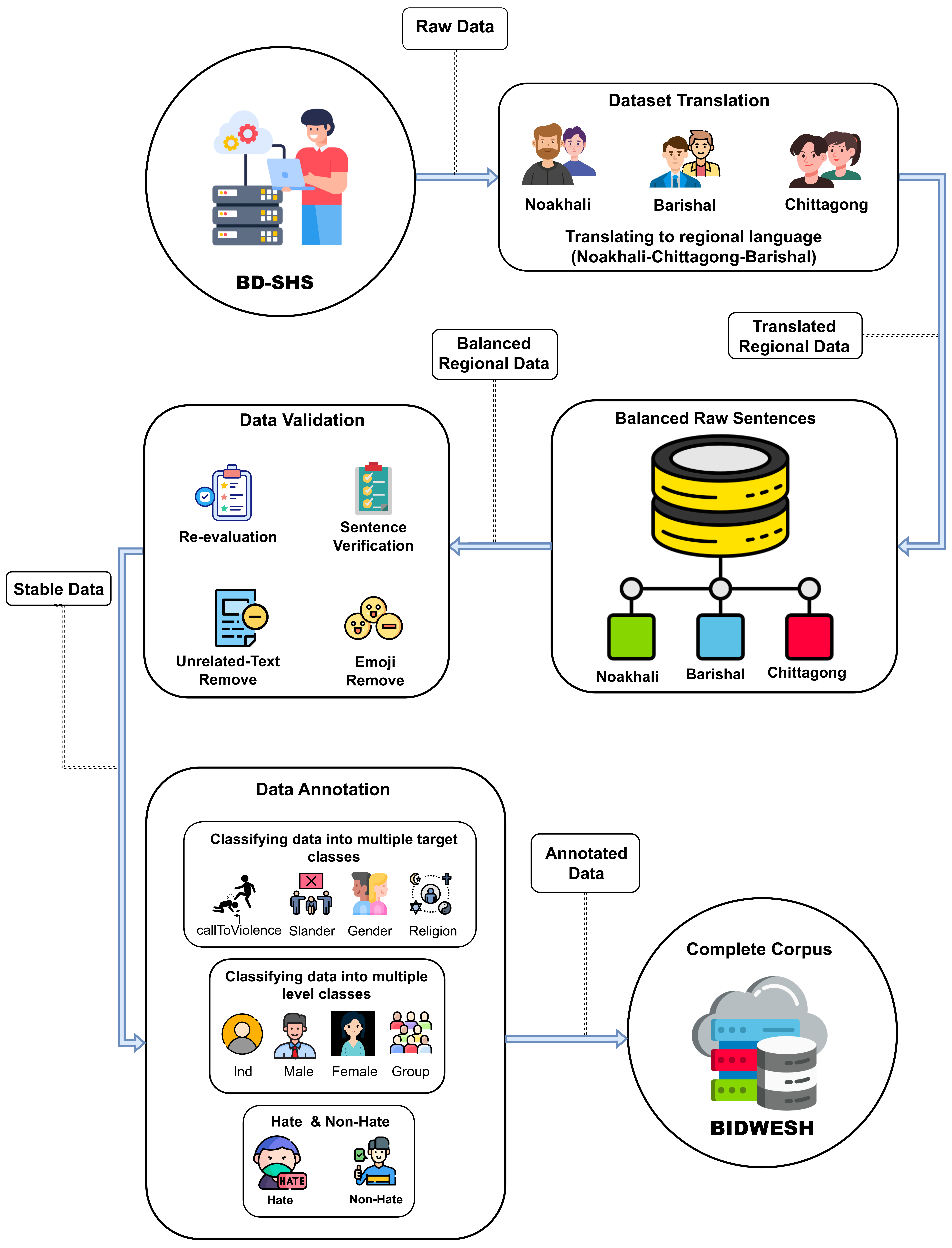}
    \caption{A Systematic Pipeline for the development of BIDWESH Dataset}
    \label{fig:Pipeline}
\end{figure}

To address this gap, we introduce BIDWESH (Hatred), the first benchmark corpus designed for hate speech detection in Bangla regional dialects. This corpus captures the diversity of dialectal speech while ensuring inclusive representation of toxic linguistic behaviors commonly used at the grassroots level. Our dataset construction process follows a comprehensive pipeline which has been illustrated in Table~\ref{fig:Pipeline}, that includes dialectal data collection, rigorous filtering, preprocessing, and expert-driven annotation. The annotation framework involved collaboration with linguistic experts proficient in regional varieties, ensuring high-quality annotations and cultural relevance.

By bridging the gap in dialectal resource availability, BIDWESH establishes the foundation for building robust and inclusive hate speech detection models capable of identifying offensive content across different Bangla-speaking communities. This work contributes a vital resource for future research in regional NLP, hate speech moderation, and computational social science.

\subsection{Data Collection}
For this study, we utilized the BD-SHS (Bangla Hate Speech) dataset as our primary data source, which is publicly available on \href{https://github.com/naurosromim/hate-speech-dataset-for-Bengali-social-media}{GitHub}. The BD-SHS dataset contains over 50,200 offensive comments originally collected from various online social networking platforms in standard Bangla. From this comprehensive dataset, we selected a subset of 3061 samples for our regional hate speech research, comprising 1513 hate speech instances and 1548 non-hate speech instances. This balanced selection ensures adequate representation of both categories for effective model training and evaluation.

\subsection{ Data Translation}

To develop a comprehensive regional hate speech corpus, we concentrated on three prominent dialectal areas of Bangladesh: Noakhali, Chittagong, and Barishal. These regions were selected based on their distinctive linguistic characteristics and significant representation in online discourse communities. The translation process involved converting over 3,061 samples into each corresponding regional variety, generating a cumulative total of more than 9,183 dialectal instances derived from the original BD-SHS collection.

The translation methodology was executed by native speakers with requisite linguistic expertise and cultural familiarity with local expressions and speech patterns. The translation framework was meticulously designed to preserve the semantic integrity and offensive characteristics of the original content while rendering it in dialectally appropriate forms that accurately represent authentic regional usage patterns. Each translation underwent collaborative review processes to resolve linguistic ambiguities, ensure inter-sample consistency, and maintain dialect-specific linguistic authenticity. All translated outputs were subsequently subjected to rigorous validation by research team members to ensure alignment with the original annotation schema of the BD-SHS dataset. This systematic and culturally-informed approach ensures that the extended corpus reflects the dialectal diversity of Bangla while providing a robust foundation for developing regionally-sensitive hate speech detection systems.

\subsection{Data Translators Identity
}
The translation process was conducted by six native speakers, comprising two translators for each regional dialect (Noakhali, Chittagong, and Barishal). All translators were native residents of their respective regions with diverse educational and professional backgrounds, including undergraduate students, software developers, and researchers. The selection criteria required all translators to be 18 years of age or older to ensure appropriateness for hate speech content translation. The cohort included two female and four male translators, selected based on translation competency evaluation and regional nativity verification by research team members. Quality assurance was maintained through systematic review processes conducted by the research team. Table~\ref{tab:annotators}, illustrates the annotators' identities.

\begin{table}[h]
\caption{Annotator Demographics}
\centering
\begin{tabular}{|l|l|l|l|l|}
\hline
\textbf{ID} & \textbf{Region} & \textbf{Background} & \textbf{Expertise} & \textbf{Age} \\
\hline
T01 & Noakhali   & Undergraduate Student & Native Speaker         & 23 \\
\hline
T02 & Noakhali   & Researcher            & Translation Experience & 25 \\
\hline
T03 & Chittagong & Undergraduate Student       & Native Speaker         & 22 \\
\hline
T04 & Chittagong & Undergraduate Student       & Native Speaker         & 23 \\
\hline
T05 & Barishal   & Undergraduate Student & Native Speaker         & 23 \\
\hline
T06 & Barishal   & Researcher    & Native Speaker         & 24 \\
\hline
\end{tabular}

\label{tab:annotators}
\end{table}

\subsection{Data Validation}
To ensure the highest quality and consistency of our dialectal hate speech dataset, we implemented a comprehensive five-stage validation process following the translation phase. This systematic approach was designed to eliminate inconsistencies, remove irrelevant content, and maintain the linguistic integrity of our regional corpus. Given that the original dataset was composed in standard Bangla and manually translated into three regional dialects—Noakhali, Chittagong, and Barishal—the risk of inconsistency was inherently high due to linguistic variability, subjective translation choices, and human error.

\subsubsection{Re-evaluation}
The initial validation stage involved a comprehensive re-evaluation conducted by the same translators who performed the original translation work. All translated sentences were manually reviewed and cross-checked against the original standard Bangla texts to verify accuracy and maintain semantic consistency. Reviewers focused on maintaining idiomatic correctness, dialect-specific vocabulary, and the offensive tone of each sentence. This process ensured that the dialectal adaptations preserved the original meaning and hate speech classification while authentically reflecting regional linguistic patterns.
\subsubsection{Unrelated Text Remove}
We systematically identified and removed unrelated textual elements that did not contribute to the core hate speech content. This included manually detecting and eliminating English text fragments and unusual textual elements that appeared inconsistent with the regional dialectal context. Additionally, we checked for entirely absent translations, partially completed sentences, and omission of key hate-bearing words or phrases. Each translated entry was validated against its original Bangla source to ensure semantic completeness and fidelity of offensive expression.
\subsubsection{Sentence Verification}
The final validation stage involved thorough sentence-level verification performed by both the original translators and additional research team members. This process encompassed grammar correction, contextual meaning verification, sentence structure optimization, and comprehensive spelling checks. Special attention was given to detecting sentences that mixed multiple dialects or included standard Bangla unintentionally, translations that were contextually inappropriate or semantically misaligned with the original sentence, and use of grammatical constructions that did not conform to the syntax of the target dialect. Due to the lack of standardized orthography for many regional Bangla dialects, spelling verification was performed manually with the aid of native speakers to ensure consistency in commonly accepted dialectal spellings and phonetic alignment with regional pronunciation.
\subsubsection{Emoji Remove}
To maintain the formal tone and textual clarity of the dataset, we performed a dedicated emoji removal step. During the translation and validation processes, some sentences retained emojis from the original standard Bangla dataset or included new emojis introduced by translators for expressive emphasis. These emojis, while potentially useful in casual communication, introduced noise and inconsistency in the textual representation of hate speech. This step ensured that the dataset remained linguistically focused and semantically uniform, without visual or symbolic artifacts that could bias downstream language models or skew hate speech interpretation.

\subsection{Data Annotation}
To prepare the BIDWESH dataset for supervised hate speech detection, a comprehensive multi-level annotation process was carried out. The annotation involved manual labeling of each sentence across multiple classification schemes to capture the full complexity of hate speech patterns in regional Bangla dialects. This systematic approach enables the development and evaluation of models capable of understanding complex offensive expression and contextual bias within Bangla regional dialects.
\subsubsection{Primary Hate Speech Classification}
Each sentence in the dataset was annotated with one of two mutually exclusive labels: Hate or Non-Hate. This binary classification serves as the primary objective for identifying offensive and toxic content within dialectal text.
\begin{itemize}
    \item \textbf{Hate}: Includes sentences containing offensive language, abusive expressions, discriminatory content, or toxic behavior directed toward individuals or groups.
    \item \textbf{Non-Hate}: Encompasses neutral content, general discussions, or non-offensive expressions without harmful intent.
\end{itemize}

\subsubsection{Multi Level Target Classification}
The annotation framework incorporates multi levels of target classification to provide a comprehensive understanding of hate speech patterns; notably, multilevel target annotations were applied exclusively to hate speech instances, with non-hate content excluded from this layer of annotation.

\textbf{Target Group Classification:}
\begin{itemize}
    \item \textbf{Individual}: Offensive content directed at a specific person, often involving personal insults or threats.  
    This includes name-calling, character attacks, or direct humiliation.
    
    \item \textbf{Male}: Hate speech that targets men, frequently using gender-specific stereotypes or derogatory terms.  
    It reflects male-directed hostility, bias, or exclusionary remarks.

    \item \textbf{Female}: Discriminatory or hostile language aimed at women, often rooted in misogyny or gender bias.  
    These instances may include objectification, stereotyping, or verbal abuse.

    \item \textbf{Group}: Aggressive or derogatory remarks directed at a community, organization, or demographic group.  
    This includes ethnic slurs, religious attacks, or regional discrimination.
\end{itemize}

\textbf{Hate Speech Type Classification:}
    \begin{itemize}
        \item \textbf{Slander}: False accusations or defamatory remarks intended to damage reputations.
        \item \textbf{Gender}: Sexist or gender-discriminatory language used to demean or harass based on gender.
        \item \textbf{Religion}: Hate speech involving religious intolerance or attacks on faith-based communities.
        \item \textbf{callToViolence}: Direct or indirect incitement to violence, aggression, or physical harm against individuals or groups.
    \end{itemize}

\subsection{Data Balancing and Stability
}
Following the annotation process, the dataset underwent careful balancing to ensure stable and representative data distribution across all regional dialects. This involved:

\begin{itemize}
    \item\textbf{Balanced Regional Data:}  Ensuring equal representation of hate and non-hate samples across Noakhali, Chittagong, and Barishal dialects.
    \item\textbf{Stable Data:}  Maintaining consistent annotation quality and distribution patterns across all demographic targets and hate speech types
    \item\textbf{Complete Corpus Formation:}  Integrating all annotated regional data into a unified, comprehensive corpus suitable for cross-dialectal hate speech detection.
\end{itemize}

\section{Data Annotation}
The annotation of the BIDWESH dataset was carried out manually by the same regional dialect experts who were previously involved in the translation process. Given their linguistic familiarity and contextual awareness, these translators were well-positioned to interpret both the literal meaning and the offensive undertone of texts within their dialectal context.

Each translator team was responsible for annotating the 3,061 dialectal sentences corresponding to their assigned region: Noakhali, Chittagong, and Barishal. They performed multi-level annotation across all classification schemes, ensuring comprehensive coverage of hate speech patterns and target demographics. The final annotated dataset represents a complete corpus with balanced regional representation, stable data distribution, and comprehensive multi-level classification suitable for training robust hate speech detection models across Bangla regional dialects. A sample of our dataset is given in Table~\ref{tab:sample}.

\begin{table}[h!]
\caption{A sample of BIDWESH dataset}
\centering
\begin{tabular}{|p{2.5cm}|p{2cm}|p{2cm}|p{2cm}|c|c|c|}
\hline
\textbf{Standard Bangla} & \textbf{Chittagong} & \textbf{Noakhali} & \textbf{Barishal} & \textbf{Target} & \textbf{Type} & \textbf{Hate Speech} \\
\hline
{\bng grur bac/ca kukuerr bac/ca.} & {\bng grur ba{I}c/ca kut/tar ba{I}c/ca.} & {\bng grur ba{I}c/cha kut/tar ba{I}c/cha.} & {\bng grur bac/ca kut/tar bac/ca.} & ind & slander & 1 \\
\hline
{\bng Aamaedr iselT Aaj ntun {I}itHas ker edkhaela.} & {\bng Aarar iselT Aaijey ntun {I}itHas gir edHa{I}iyY.} & {\bng Aanega iselT Aa{I}j/ja enaJa {I}itHas kir edkha{I}es.} & {\bng emaega iselT Aa{I}j ntun {I}itHas Her edHa{I}ela.} & - & - & 0 \\
\hline
\end{tabular}

\label{tab:sample}
\end{table}

\section{Data Statistics}
The BIDWESH dataset represents a comprehensive corpus for hate speech detection in Bangla regional languages, comprising 9,183 instances systematically collected from three distinct regional dialects of Bangladesh. This dataset was meticulously developed through manual translation from the BD-SHS dataset to preserve the linguistic nuances and cultural context inherent in regional Bangla variations. The corpus is designed to address the critical gap in hate speech detection research for regional Bangla languages, providing a balanced representation across geographical and linguistic boundaries.
\subsection{Data Distribution}
\begin{figure}[H]
    \centering
    \includegraphics[width=0.8\textwidth]{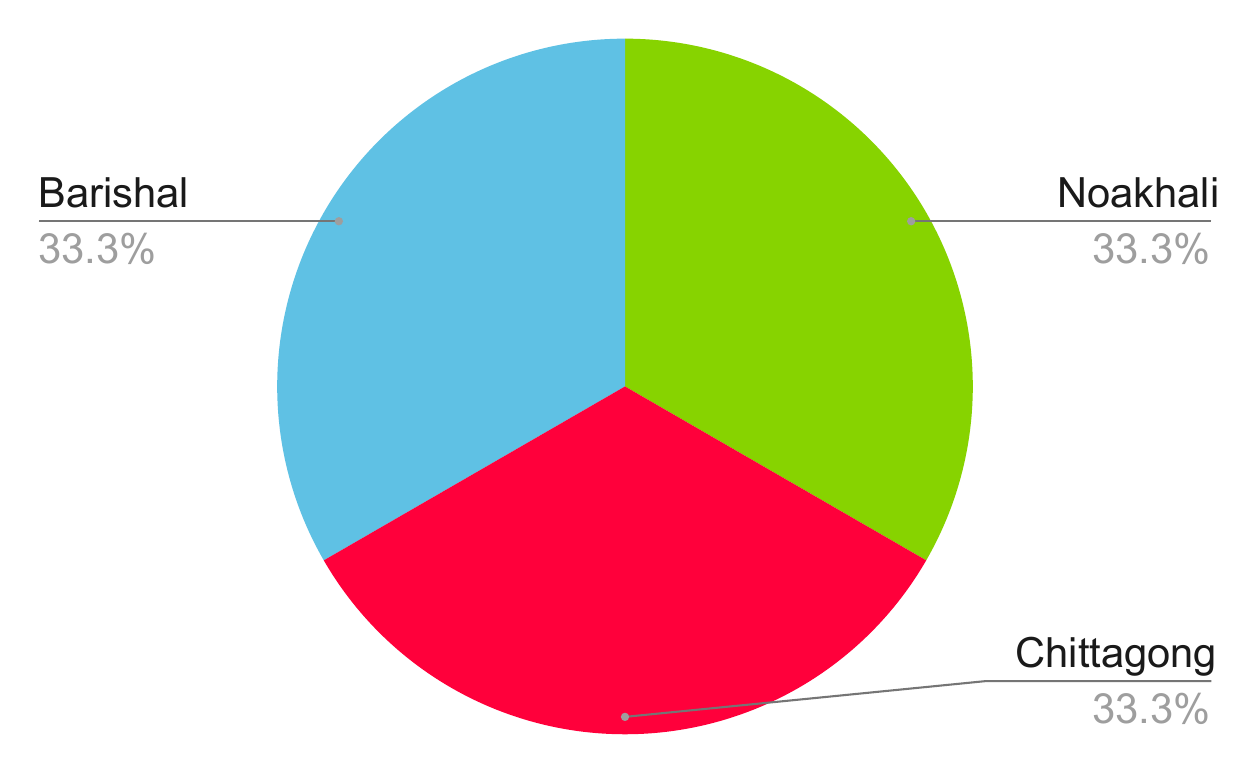}
    \caption{Regional Distribution of Dialectal Instances}
    \label{fig:diversity in Bangla dialects}
\end{figure}

The dataset maintains perfect regional balance, with each of the three major dialectal regions contributing equally to ensure comprehensive linguistic coverage: Figure~\ref{fig:diversity in Bangla dialects} illustrates the distribution of samples in each
Bangla dialects.

\subsection{Hate Speech Classification}
In our dataset, hate speech instances are labeled with the value 1, while non-hate speech instances are labeled with the value 0. Additionally, we classified two more classes: Type and Target classes. These classes are only applicable to hate speech instances. The binary hate speech classification demonstrates a well-balanced distribution, essential for robust model training and evaluation: Table~\ref{tab:class_distribution} illustrates the Class Distribution of Instances.
\begin{table}[H]
\caption{Class Distribution of Instances}
\centering
\begin{tabular}{|l|c|r|r|}
\hline
\textbf{Class} & \textbf{Label} & \textbf{Instances} & \textbf{Percentage} \\
\hline
Hate Speech     & 1 & 1,513 & 49.24\% \\
\hline
Non-Hate Speech & 0 & 1,548 & 50.76\% \\
\hline
\textbf{Total}  &    & \textbf{3,061} & \textbf{100\%} \\
\hline
\end{tabular}
\label{tab:class_distribution}
\end{table}
\subsection{Type Classification}
\begin{figure}[H]
    \centering
    \includegraphics[width=0.8\textwidth]{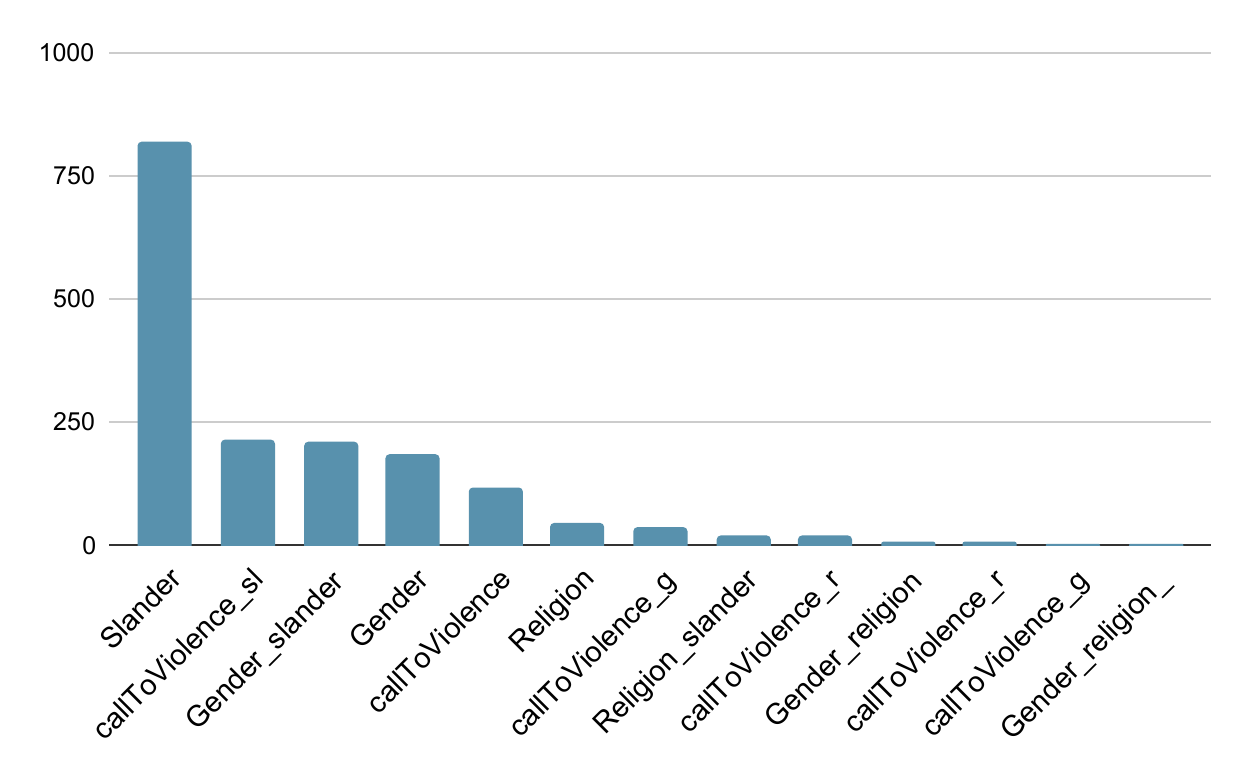}
    \caption{Type Classification}
    \label{fig:Type Classification}
\end{figure}

The type classification encompasses 13 distinct categories, capturing the multifaceted nature of hate speech manifestations. The distribution reveals the complexity of hate speech patterns, with both single-label and multi-label combinations. Table~\ref{tab:type_category_distribution} and Figure~\ref{fig:Type Classification} illustrates the Regional Distribution of Dialectal Instances.
\begin{table}[h]
\caption{Distribution of Hate Speech Type Categories}
\centering
\begin{tabular}{|l|r|r|}
\hline
\textbf{Type Category} & \textbf{Instances} & \textbf{Percentage} \\
\hline
Slander                                         & 822  & 53.65\% \\
\hline
callToViolence\_slander                         & 214  & 13.96\% \\
\hline
Gender\_slander                                 & 210  & 13.71\% \\
\hline
Gender                                          & 186  & 12.14\% \\
\hline
callToViolence                                  & 119  & 7.77\%  \\
\hline
Religion                                        & 47   & 3.07\%  \\
\hline
callToViolence\_gender                          & 36   & 2.35\%  \\
\hline
Religion\_slander                               & 23   & 1.50\%  \\
\hline
callToViolence\_religion                        & 22   & 1.44\%  \\
\hline
Gender\_religion                                & 8    & 0.52\%  \\
\hline
callToViolence\_religion\_slander               & 8    & 0.52\%  \\
\hline
callToViolence\_gender\_slander                 & 4    & 0.26\%  \\
\hline
Gender\_religion\_slander                       & 3    & 0.20\%  \\
\hline
\textbf{Total}                                  & \textbf{1,513} & \textbf{100\%} \\
\hline
\end{tabular}
\label{tab:type_category_distribution}
\end{table}

\subsection{Target Classification}
\begin{figure}[H]
    \centering
    \includegraphics[width=0.8\textwidth]{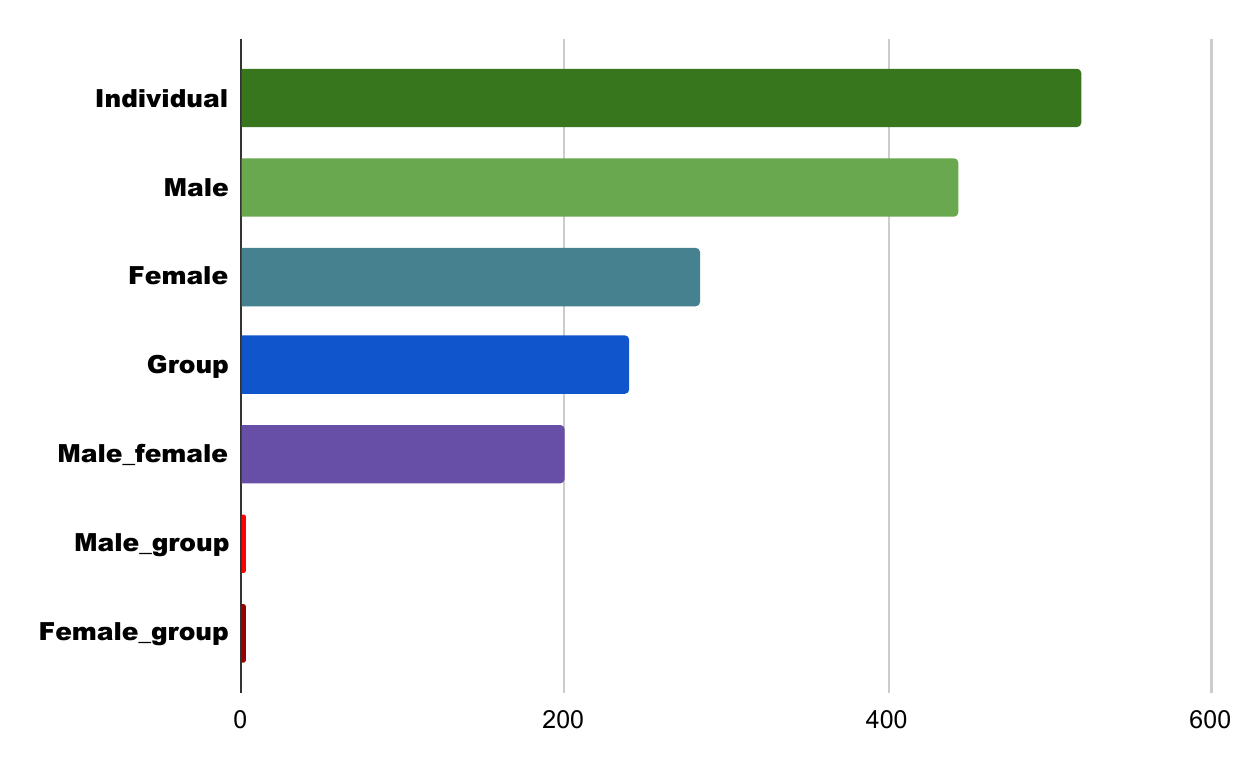}
    \caption{Target Classification}
    \label{fig:Target Classification}
\end{figure}

The target classification identifies seven distinct categories, representing the diverse targeting mechanisms employed in hate speech across the three regional dialects: Table~\ref{tab:target_category} and Figure~\ref{fig:Target Classification} illustrates the Regional Distribution of Dialectal Instances.
\begin{table}[h]
\caption{Target Category Distribution}
\centering
\begin{tabular}{|l|c|c|}
\hline
\textbf{Target Category} & \textbf{Instances} & \textbf{Percentage} \\
\hline
Individual (ind) & 520 & 34.37\% \\
\hline
Male & 443 & 29.28\% \\
\hline
Female & 284 & 18.77\% \\
\hline
Group & 240 & 15.87\% \\
\hline
Male\_female & 200 & 13.22\% \\
\hline
Male\_group & 3 & 0.20\% \\
\hline
Female\_group & 3 & 0.20\% \\
\hline
\textbf{Total} & \textbf{1513} & \textbf{100\%} \\
\hline
\end{tabular}
\label{tab:target_category}
\end{table}

\section{Dataset Characteristics}
The BIDWESH dataset exhibits several key characteristics that distinguish it as a valuable resource for hate speech detection research in regional Bangla languages:
\begin{itemize}
    \item \textbf{Regional Diversity:} Equal representation from Barishal, Noakhali, and Chattogram dialects ensures comprehensive linguistic coverage.
    
    \item \textbf{Balanced Classification:} Near-equal distribution of hate (49.24\%) and non-hate (50.76\%) data enables unbiased model training and evaluation.
    
    \item \textbf{Multi-dimensional Annotation:} Three-level annotation—hate speech presence, type, and target—enables detailed and layered analysis.
    
    \item \textbf{Complex Type Taxonomy:} 13 distinct hate type categories, including multi-label cases, capture the nuanced and overlapping nature of hate speech.
    
    \item \textbf{Comprehensive Target Coverage:} Seven target classes reflect a wide range of hate targets, from individuals to gender and group-based categories.
    
    \item \textbf{Manual Translation:} Regional authenticity is preserved through manual translation of BD-SHS data into local dialects.
\end{itemize}

\section{Ethics Statements}
The data used to construct BIDWESH does not raise ethical concerns, as it was collected from a publicly available dataset and through manual translation. The dataset contains no sensitive or private information, ensuring compliance with ethical standards. Additionally, no data was sourced from personal communications or restricted sources. This dataset is intended solely for NLP research and development; no human or animal subjects were involved in its creation.

\section{Availability of BIDWESH}
The BIDWESH dataset, containing annotated sentiment analysis data in Bangla regional dialects, is publicly available for research and academic purposes. Researchers interested in utilizing the dataset can access it through the following link:\href{https://data.mendeley.com/datasets/bpkrvf882k/1}{Dataset of Hate Speech Detection for Regional Bangla Language (Original data)} (Mendeley Data)

\section{CRediT Author Statement}

 \textbf{Azizul Hakim Fayaz:} Conceptualization, Data curation, Methodology, Writing – original draft.\\ 
 \textbf{MD. Shorif Uddin:} Conceptualization, Formal analysis, Investigation, Methodology, Writing – original draft. \\
 \textbf{Rayhan Uddin Bhuiyan:} Data curation, Writing – original draft. \\ 
 \textbf{Zakia Sultana:} Data curation, Formal analysis, Methodology. \\
 \textbf{Md. Samiul Islam:} Investigation, Validation, Writing – review \& editing. \\ 
 \textbf{Bidyarthi Paul:} Supervision, Writing – review \& editing. \\
 \textbf{Tashreef Muhammad:} Validation. \\ 
 \textbf{Shahriar Manzoor:} Supervision. \\



\bibliography{references}
\bibliographystyle{unsrt}  
\end{document}